\documentclass[conference]{IEEEtran}
\IEEEoverridecommandlockouts
\usepackage{cite}
\usepackage{amsmath,amssymb,amsfonts}
\usepackage{algorithmic}
\usepackage{graphicx}
\usepackage{textcomp}
\usepackage{xcolor}
\usepackage{amsmath,amssymb}
\usepackage{multirow}
\usepackage{comment}
\def\BibTeX{{\rm B\kern-.05em{\sc i\kern-.025em b}\kern-.08em
    T\kern-.1667em\lower.7ex\hbox{E}\kern-.125emX}}
\begin{document}

\title{Class Equilibrium using Coulomb's Law}

\author{
\IEEEauthorblockN{Saheb Chhabra$^{1}$, Puspita Majumdar$^{1}$, Mayank Vatsa$^{2}$, and Richa Singh$^{2}$}
\IEEEauthorblockA{$^{1}$IIIT-Delhi, India; $^{2}$IIT Jodhpur, India\\Email: \{sahebc, pushpitam\}@iiitd.ac.in; \{mvatsa, richa\}@iitj.ac.in
}
}

\maketitle

\begin{abstract}
Projection algorithms learn a transformation function to project the data from input space to the feature space, with the objective of increasing the inter-class distance. However, increasing the inter-class distance can affect the intra-class distance. Maintaining an optimal inter-class separation among the classes without affecting the intra-class distance of the data distribution is a challenging task. In this paper, inspired by the Coulomb's law of Electrostatics, we propose a new algorithm to compute the equilibrium space of any data distribution where the separation among the classes is optimal. The algorithm further learns the transformation between the input space and equilibrium space to perform classification in the equilibrium space. The performance of the proposed algorithm is evaluated on four publicly available datasets at three different resolutions. It is observed that the proposed algorithm performs well for low-resolution images.
\end{abstract}


\section{Introduction}
Increasing the inter-class distance is at the heart of the majority of loss functions and projection algorithms designed in the machine learning community. However, as the Newton's Third Law of Motion states that for every action, there is an equal and opposite reaction; increasing the inter-class distance often affects the intra-class distance as well. In both the cases, maintaining equilibrium between the two forces or the two criterion functions is an important challenge. 

\begin{figure}[t]
\centering
\includegraphics[scale = 0.85]{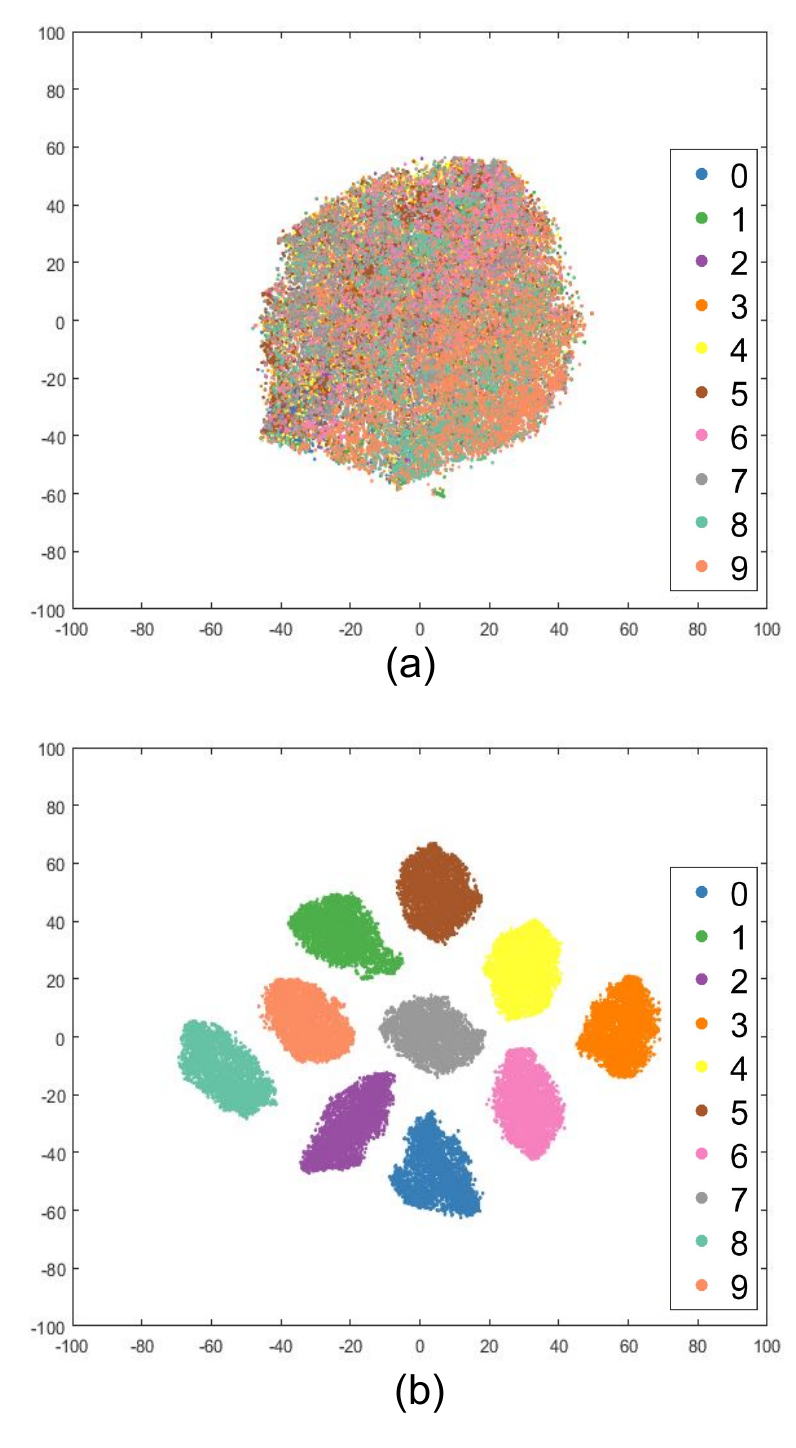}

\caption{Illustration of class equilibrium on the CIFAR-10 dataset. t-SNE visualization of training data points in (a) input space and (b) equilibrium space. In input space, the classes are overlapped while in equilibrium space, the separation among the classes is optimal}
\label{fig:Intro_CIFAR_Image}
\end{figure}

In machine learning, the projection of data into feature space plays an important role in improving the performance of the classifier. Several algorithms such as manifold learning, metric learning, and deep neural networks have been proposed to transform the data from the input space to a different feature space where the classes are separable \cite{wold1987principal,martinez2001pca,schroff2015facenet,shi2003shrinking, singhPAMI}. However, existing data projection techniques depend on the training data distribution and may get stuck in local optima. Essentially, the projected space is dependent on the choice of projection algorithm. In this research, we propose an alternative perspective to it by asking the question: ``\textit{What will be the optimal projection space for this data distribution}"? The question can be answered by computing a space where the competing influences of all the classes are balanced.

The study of data driven projection algorithms suggest that a more fundamental approach which balances the forces exerted by samples of multiple classes could provide generalizable results. Inspired by different laws of Physics, researchers have proposed different techniques for data projection, classification, and clustering \cite{qian2017hybrid, kushwaha2019electromagnetic, li2019improved, zhang2020novel, li2020hibog, fister2020potential, sureka2020nature, unterthiner2017coulomb}. Inspired by Newton's universal law of gravitation, Shi \textit{et al.} \cite{shi2003shrinking} presented a data preprocessing technique to optimize the inner structure of the data to create condensed and widely separated clusters. Hochreiter \textit{et al.} \cite{hochreiter2003coulomb} introduced a family of classifiers called Coulomb classifiers, that include v-SVM and C-SVM. Coulomb classifier is based on physical analogy to an electrostatic system of charged conductors. It is shown that Coulomb classifier can outperform a standard SVM. Schiegg \textit{et al.} \cite{schiegg2016learning} proposed Coulomb Structured SVM (CSSVM) to get an ensemble of models at training time that can run in parallel and independently to make a diverse prediction during testing time. Another application of Law of Physics in classification algorithm is shown in the work by Shafigh \textit{et al.} \cite{shafigh2013gravitation}. In this work, the gravitational potential energy between particles is used to develop a new classification method. The aim of this algorithm is to find the equilibrium condition of the classifier. Inspired by electric network theory, \textit{Hirai et al.} \cite{hirai2007electric} proposed a new classifier for semi-supervised learning on graphs. The classifier classifies the dataset using the sign of electric potential. This classifier can handle complex real word problems and give better performance than diffusion kernel methods. Peng and Liu \cite{peng2017gravitation} proposed a multi-label classification algorithm termed as ITDGM. The proposed algorithm models the given problem using gravitational theory and uses the interaction-based gravitation coefficient to compute the gravitational force for classification. A universal gravity rule based clustering algorithm is proposed by Alswaitti et al. \cite{alswaitti2018optimized}. In this work authors proposed different solutions to impose a balance between exploitation and exploration. Vashishtha et al. \cite{vashishtha2018novel} proposed a novel framework that discovers the classification rules which can be used with the combination of any nature inspired algorithms. Recently, Wang et al. \cite{wang2020entropy} proposed a meta heuristic Nearest Neighbor (NN) algorithm to overcome the drawbacks in NN-based classifiers for imbalanced data problems. The proposed algorithm termed as Gravitational Fixed Radius Nearest Neighbor classifier (GFRNN), solves imbalanced problems by drawing on Newton’s law of universal gravitation.
From the existing literature, it is observed that Laws of Physics used for data projection and classification lead to good result and can be applied to real-world problems. However, the two important challenges in computing an optimal projection space for a data distribution are be formulated as:
\begin{enumerate}
\item Computation of the projection space should be independent of the complexity of data distribution.
\item The separation among the classes should be optimal in the projection space.
\end{enumerate}

In this research, we propose a novel algorithm that uses Coulomb's Law of repulsion in Electrostatics to transform the data from input space to an equilibrium space where the separation among the classes is optimal. The intra-class distance between the data points of each class in equilibrium space remains the same and the position of each class with respect to another class is optimal. Transformation function between input space and equilibrium space is learned to perform classification in equilibrium space.



\begin{figure*}[t]
\centering
\includegraphics[scale = 0.35]{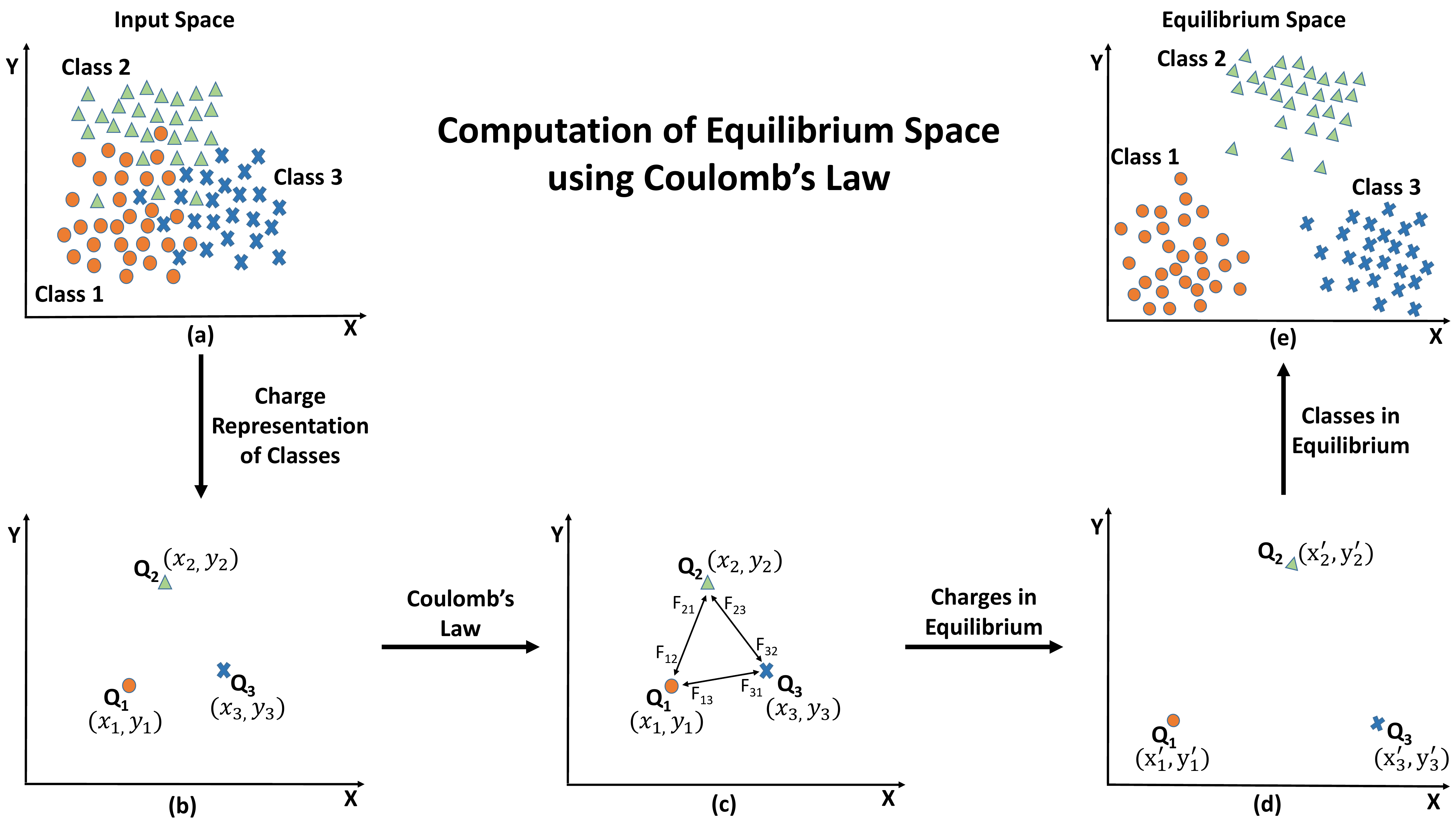}
\caption{Block diagram of the proposed algorithm for computation of the equilibrium space (a) representation of class distribution in input space, (b) shows the charge representation of the classes, (c) illustrates the force applied by the charges using Coulomb's law on each other to move in equilibrium space, (d) shows the charges in equilibrium space, and (e) represents the classes in equilibrium space.}
\label{fig:Block_Diagram}
\end{figure*}

\section{Coulomb's Law and Class Equilibrium}

Coulomb's law \cite{jackson1999classical} describes the force interacting between electric charges. Electric charge causes an object to experience a force when placed in an electromagnetic field. It is of two types: 1) positive charge carried by proton and 2) negative charge carried by electron. From the theory of electrostatics, it is known that opposite charge attracts and same charge repels. So if there are two point charges with charge $Q_1$ and $Q_2$ respectively in space, then Coulomb's law states that ``The magnitude of the electrostatic force of attraction or repulsion between two point charges is directly proportional to the product of the magnitudes of charges and inversely proportional to the square of the distance between them''. Mathematically, it is written as:
\begin{equation}
F = k\frac{|Q_1| |Q_2|}{|R|^2}
\end{equation}
where, $k$ is a constant, $F$ is the magnitude of electrostatic force between charges $Q_1$ and $Q_2$, $R$ is the distance between the point charges. If $\mathbf{R_1}$, $\mathbf{R_2}$ are the position vectors of charge $Q_1$ and $Q_2$ respectively then the vector form of Coulomb's law can be written as:
\begin{equation}
\mathbf{F_{12}} = k \frac{Q_1 Q_2}{|\mathbf{R_{21}}|^2} \hat{\mathbf{R_{21}}}, \hspace{15pt}
 \label{eq:Vector_Form}
\mathbf{F_{12}} = k \frac{Q_1 Q_2}{|\mathbf{R_2} - \mathbf{R_1}|^3} (\mathbf{R_2} - \mathbf{R_1})
\end{equation}
where, $\mathbf{F_{12}}$ represents the force on charge $Q_1$ due to charge $Q_2$. In case if multiple charges are present at the same time in space, the total force on a particular charge $Q_0$ is the vector sum of the individual forces. Mathematically, it is written as:
\begin{equation}
\label{eq:Force_vector}
\mathbf{F_0} = \sum_{i} \mathbf{F_{0i}} \quad \forall i, \; i \neq 0
\end{equation}
where, $\mathbf{F_0}$ represents the vector sum of the forces acting on charge $Q_0$ and $\mathbf{F_{0i}}$ represent the force by charge $Q_i$ on charge $Q_0$.

Equilibrium of a system is defined as the state of the system where the overall force on each class with respect to other classes is balanced. In the proposed algorithm, Coulomb's law is applied to the data points in input space to achieve class equilibrium. Following are the properties of class equilibrium:

\begin{enumerate}
\item the separation among the classes in equilibrium space is optimal,
\item the overall force on the classes in equilibrium is close to zero.
\end{enumerate}

Sample t-SNE visualization of the input and equilibrium space is demonstrated on the CIFAR-10 training images in Fig. \ref{fig:Intro_CIFAR_Image}. It is our assertion that data projection into this space will significantly improve the classification performance.  

\section{Proposed Approach} 

The problem statement can be formally defined as ``If there are $n$ number of classes in input space $\chi$ then project the data points from input space $\chi$ to an equilibrium space $E$ where the separation among the classes is optimal. After computing the equilibrium space, learn the transformation between input space and equilibrium space to project the data points to equilibrium space and perform classification". Inspired by Coulomb's Law, this research propose a new algorithm to compute an equilibrium space where the classes are optimally separated so that the classification can be performed in equilibrium space. Fig. \ref{fig:Block_Diagram} shows the block diagram of the proposed algorithm for computing equilibrium space using training data. 

Let us assume that there are three classes in the input space and we want to compute the equilibrium space as shown in Fig. \ref{fig:Block_Diagram}. In the \textbf{first step}, each class is represented with a positive charge of magnitude $Q_1$, $Q_2$, and $Q_3$ at positions $(x_1, y_1)$, $(x_2, y_2)$, and $(x_3, y_3)$, respectively in the input space. In the \textbf{second step}, Coulomb’s law of repulsion is applied on the system of charges such that the charges move and settle at positions $(x_1', y_1')$, $(x_2', y_2')$, and $(x_3', y_3')$, respectively in the equilibrium space; where the net force of the system is close to zero. In the \textbf{final step}, the data points of each class are shifted with respect to the position of the charges in equilibrium space. This results in the optimal projection of the classes in equilibrium space. Once the input data is projected in the equilibrium space, the next task is to learn the transformation between input space and equilibrium space and perform classification. For this purpose, convolutional neural network (CNN) is used to learn the transformation function using the training data. The learned transformation may result in some error and thus an error removal technique (ERC) (discussed in subsection \ref{classification_eq}) is used. For classification, a negative charge of unit magnitude is assigned to each test data point. After that, a test point is projected in the equilibrium space using the learned transformation. Each class in equilibrium space will attract the test point towards itself. Finally, the test data point will be assigned to the class that exerts maximum force. The problem is divided into two phases: 1) computation of equilibrium space and 2) classification in equilibrium space. The detailed description of the two phases is discussed next.

\subsection{Computation of Equilibrium Space}

\label{Eq_Space}

Let $\mathbf{C}^{\chi}$ be the class set with $n$ number of classes where each class $\mathbf{C}_i^{\chi}$ has $m_i$ number of data points in the input space $\mathbf{\chi}$. Mathematically, it is represented as:
\begin{equation}
\label{OrigTrainSet}
    \mathbf{C}^{\chi} = \{\mathbf{C}_1^{\chi}(m_1), \mathbf{C}_2^{\chi}(m_2),...\mathbf{C}_n^{\chi}(m_n)\}
\end{equation}

The first task is to represent each class $\mathbf{C}_i^{\chi}$ with charge $Q_i$ at position $\mathbf{R}_{Q_i, \chi}$ as shown in Fig. \ref{fig:Block_Diagram}(b). Mathematically, it is represented as:

\begin{equation}
\label{Charge_position_Equation}
Q_i = f_1(\mathbf{C}_i^{\chi}) \quad \text{and} \quad \mathbf{R}_{Q_i, \chi} = f_2(\mathbf{C}_i^{\chi})
\end{equation}

where, function $f_1(\mathbf{C}_i^{\chi})$ outputs the charge $Q_i$ corresponding to class $\mathbf{C}_i^{\chi}$ and function $f_2(\mathbf{C}_i^{\chi})$ outputs the position vector $\mathbf{R}_{Q_i, \chi}$ of the charge $Q_i$ in input space $\chi$. Let $\mathbf{Q}$ represents the set with $n$ charges obtained corresponding to $n$ classes and $\mathbf{R}_{\mathbf{Q}, \chi}$ represent the position vector of $\mathbf{Q}$ in the input space. 
The next task is to map the charge set $\mathbf{Q}$ with position vector $\mathbf{R}_{\mathbf{Q}, \chi}$ from input space $\chi$ to the equilibrium space $E$ as shown in Fig. \ref{fig:Block_Diagram}(b),  \ref{fig:Block_Diagram}(d). 
\begin{equation}
\label{Train_Eq_Equation}
\mathbf{R}_{\mathbf{Q}, E} = z(\mathbf{Q}, \mathbf{R}_{\mathbf{Q}, \chi})
\end{equation}

where, $\mathbf{R}_{\mathbf{Q}, E}$ represents the position vector of $\mathbf{Q}$ in the equilibrium space $E$. $z(\mathbf{Q},  \mathbf{R}_{\mathbf{Q}, \chi})$ is the function that maps set $\mathbf{Q}$ with position vector $\mathbf{R}_{\mathbf{Q}, \chi}$ in the input space to the equilibrium space $E$. 

After obtaining the position vector of $n$ charges in equilibrium space, the next task is to project the data points of each class $\mathbf{C}_i^{\chi}$ in input space to equilibrium space as shown in Fig. \ref{fig:Block_Diagram}(e). Data points of each class $\mathbf{C}_i^{\chi}$ are therefore shifted by adding a vector $\mathbf{\delta}_i$ corresponding to each class. The vector $\mathbf{\delta}_i$ is computed as: 
\begin{equation}
\label{eq:Shift_Equation}
\mathbf{\delta}_i = \mathbf{R}_{Q_i, E} - \mathbf{R}_{Q_i, \chi}
\end{equation}

In order to solve Equations \ref{Charge_position_Equation} and \ref{Train_Eq_Equation}, the problem is divided into two stages. 

\noindent \textbf{Stage 1 - Charge and Position Representation}: To compute the position vector $\mathbf{R}_{Q_i, \chi}$ corresponding to each class $\mathbf{C}_i^\chi$ in input space $\chi$, mean of each class is computed.
\begin{equation}
\mathbf{R}_{Q_i, \chi} = \frac{1}{m_i}\sum \mathbf{C}_i^\chi(m_i)
\end{equation}
The charge $Q_i$ is computed by taking the spread of each class $\mathbf{C}_i^\chi$. 
\begin{equation}
Q_i = var(\mathbf{C}_i^{\chi})
\end{equation}
\begin{figure}[t]
\centering
\includegraphics[scale = 0.32]{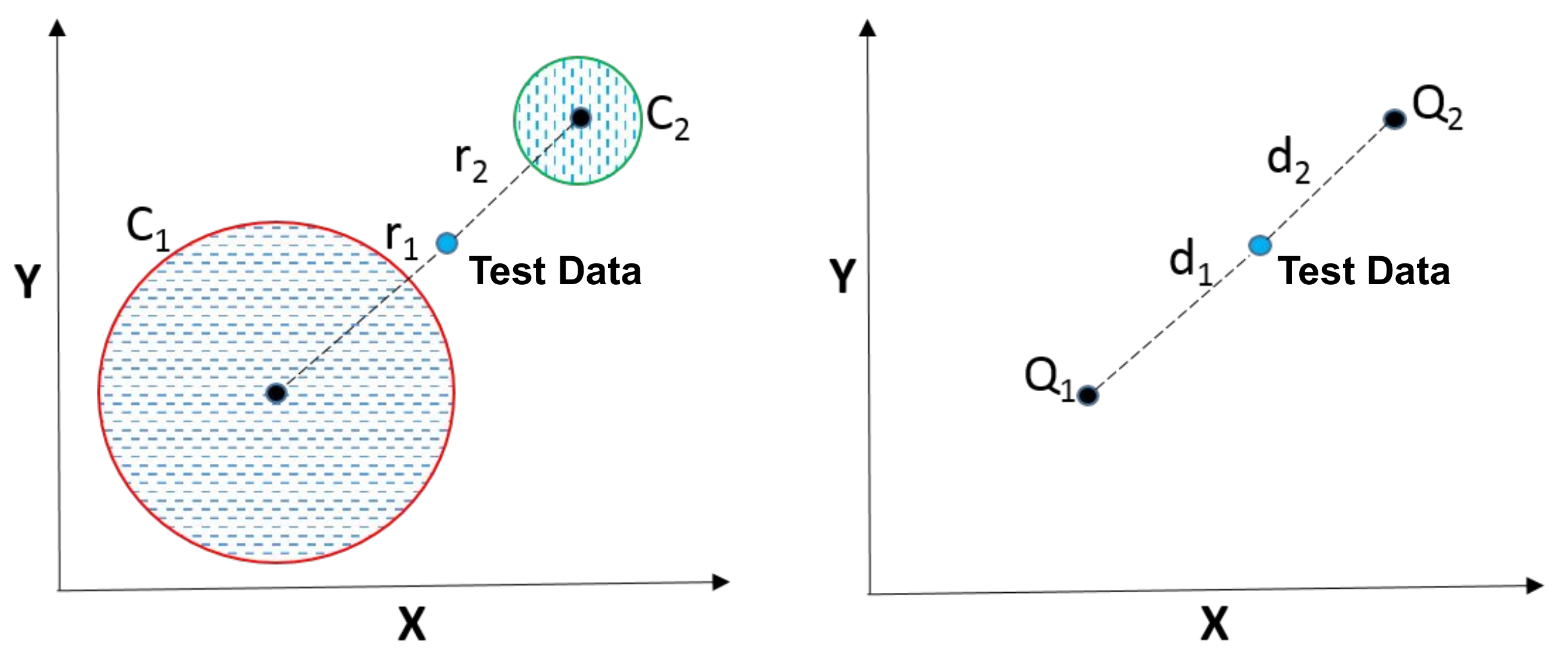}
\caption{Example to illustrate the purpose for taking variance as a metric for computing charge of each class. }
\label{fig:stage-1}
\end{figure}
The reason for taking "variance" as the metric for computing charge $Q_i$ is discussed below : 

\textbf{Analogy:} As shown in Fig. \ref{fig:stage-1}, the test data point is close to the mean of class $C_2$ as compared to class $C_1$. However, intuitively the test data point should belong to class $C_1$. On comparing this with the electrostatic theory, if we have two positive charges with magnitude $Q_1$ and $Q_2$ at the mean location of class $C_1$ and $C_2$ respectively, and another negative charge of unit magnitude at the location of the test data point, then the charge with greater magnitude will attract the test data point more towards itself as compared to the charge with smaller magnitude. From Fig. \ref{fig:stage-1}, it can be observed that the spread of class $C_1$ is more than class $C_2$ and thus has more influence on the test data point. Therefore, representing charge with the class variance is necessary and serves as a good metric.

\noindent \textbf{Stage 2 - Equilibrium Space Computation}: At this stage, the task is to compute the equilibrium space where the projection of the classes is optimal. Here, optimality is achieved when the overall force of the system in equilibrium space is close to zero. 

\begin{figure}[t]
\centering
\includegraphics[scale = 0.35]{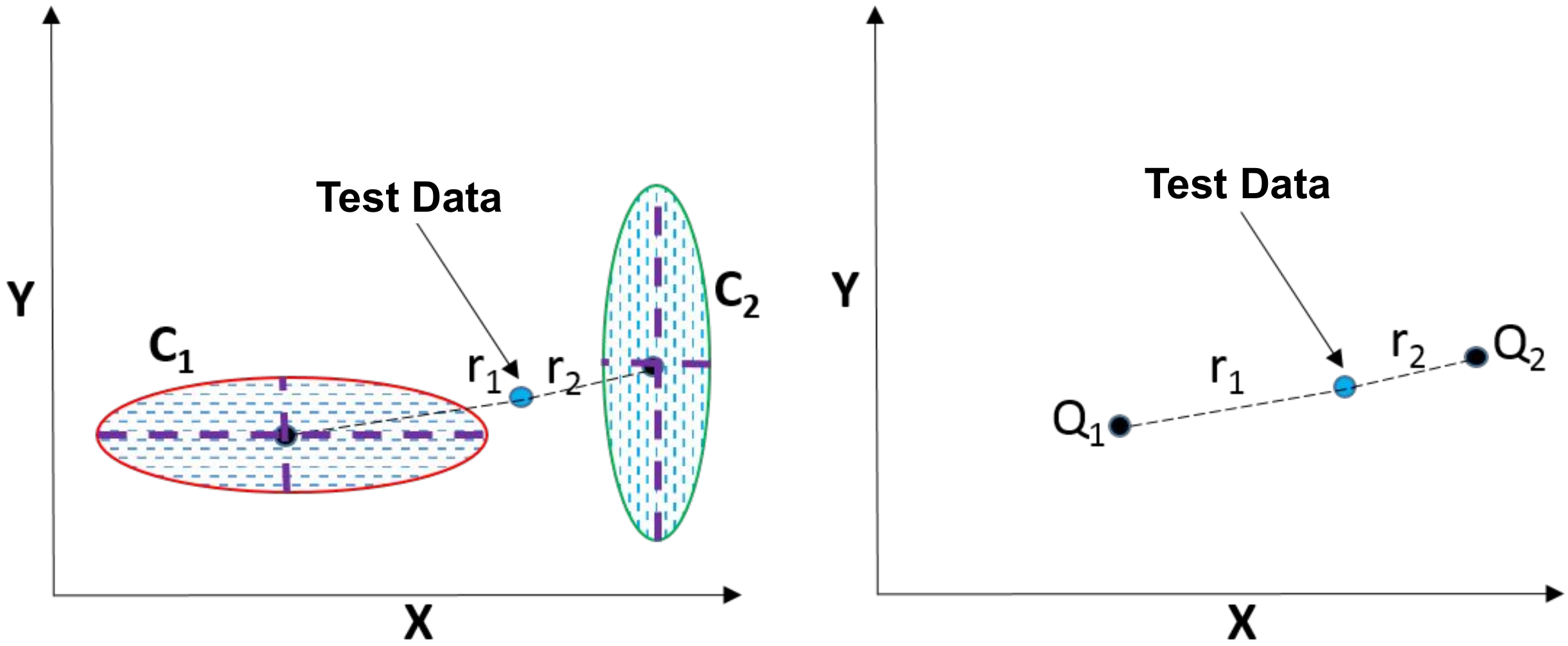}
\caption{Example to illustrate the purpose of providing variance in each direction as a weight while computing force. In this example, the variance in dimension $x$ and $y$ is different for both the classes. Class $C_1$ has greater variance in $x$ direction while class $C_2$ has in $y$ direction. Therefore, the charge representing class $C_1$ should have greater force in $x$ direction and the charge representing class $C_2$ should have greater force in $y$ direction. In order to provide weight to the force direction, variance corresponding to each dimension is multiplied with the force vector.}
\label{fig:Case-2}
\end{figure}

Once the charge $Q_i$ with position vector $\mathbf{R}_{Q_i, \chi}$ corresponding to each class $\mathbf{C}_i^\chi$ is obtained, the next task is to project the charges from input space $\chi$ to equilibrium space $E$ to obtain the position vector $\mathbf{R}_{\mathbf{Q}, E}$. For this purpose, Coulomb's law is applied such that the overall force of the system in equilibrium space is close to zero. The optimization function is given by:
\begin{equation}
\mathbf{R}_{\mathbf{Q}, E} = \underset{\mathbf{R}_{\mathbf{Q}, \chi}}{\arg\min} \sum_{i = 1}^{n} |\mathbf{F}_i|
\end{equation}
$$ \text{subject to} \sum |\mathbf{F}_i| \approx 0$$
where, $|\mathbf{F}_i|$ represents the magnitude of the force.
Using Coulomb's Force Equation \ref{eq:Vector_Form}, the above equation can be written as:
\begin{multline}
\label{eq:optimization_1}
\mathbf{R}_{\mathbf{Q}, E} = \underset{\mathbf{R}_{\mathbf{Q}, \chi}}{\arg\min} \\ \sum_{i = 1}^{n} \left|\sum_{j = 1}^{n} k\frac{Q_i Q_j}{|\mathbf{R}_{Q_j, \chi} - \mathbf{R}_{Q_i, \chi}|^3 } (\mathbf{R}_{Q_j, \chi} - \mathbf{R}_{Q_i, \chi})\right| \quad i \neq j
\end{multline}
such that $\sum |\mathbf{F}_i| \approx 0$. It is possible that the spread of each class $\mathbf{C}_i^\chi$ is not uniform in each dimension as shown in Fig. \ref{fig:Case-2}. So the force applied by a charge should be different in each dimension depending upon the spread of the data. Therefore, the force applied by a charge $Q_i$ is a function of the spread $\mathbf{S}_i$ of the data, where, $\mathbf{S}_i$ is a $d$ dimensional vector representing the variance in each dimension. Thus the force is weighted with the spread $\mathbf{S}_i$. Therefore, the Equation \ref{eq:optimization_1} is updated as:
\begin{multline}
\mathbf{R}_{\mathbf{Q}, E} = \underset{\mathbf{R}_{\mathbf{Q}, \chi}}{\arg\min} \\ \sum_{i = 1}^{n} \left|\sum_{j = 1}^{n} k\frac{Q_i Q_j}{|\mathbf{R}_{Q_j, \chi} - \mathbf{R}_{Q_i, \chi}|^3 } (\mathbf{R}_{Q_j, \chi} - \mathbf{R}_{Q_i, \chi}) \mathbf{S}_j\right| \quad i \neq j
\end{multline}

The above equation is optimized over position vector $\mathbf{R}_{Q_i, \chi}$. After optimization, the position vector of each charge $Q_i$ in equilibrium space is obtained.


Once the position vector in equilibrium space is obtained, the shift in position vector of each charge corresponding to each class from input space to equilibrium space is computed using Equation \ref{eq:Shift_Equation}. After that each class $\mathbf{C}_i^\chi$ is projected to the equilibrium space by adding the shift vector $\mathbf{\delta}_i$ with the data points of each class. Mathematically, it is written as:
\begin{equation}
\mathbf{C}_i^E = \mathbf{C}_i^\chi + \mathbf{\delta}_i
\end{equation}

\subsection{Classification in Equilibrium Space}
\label{classification_eq}
The classification in equilibrium space is divided into stages: 1)  Transformation learning using convolutional neural network (CNN) 2) Coulomb classification. The description of the stages is given below:

\noindent \textbf{Stage 1: Transformation Learning using CNN}

Once the points in input and equilibrium space are obtained, the next task is to learn the transformation function between the points $\mathbf{C}_i^\chi$ in input space $\chi$ and points $\mathbf{C}_i^E$ in equilibrium space $E$. For this purpose, a convolutional neural network (CNN) is learned. Mathematically, it is represented as:
\begin{equation}
\mathbf{C}_i^E = \phi (\mathbf{C}_i^\chi,\mathbf{W},b)
\end{equation}
where, $\phi$ is the CNN, $\mathbf{W}$ and $b$ represents the weight and bias, respectively. Mean square error loss function is used to minimize the error between the data points $\mathbf{C}_i^\chi$ in input space $\chi$ and the data points $\mathbf{C}_i^E$ in equilibrium space $E$.

\textbf{ERC - Error Removal using Correlation:}
Using the learned transformation, data points are transformed from input space $\chi$ to equilibrium space $E$. However, it is important to note that the mapping from input space to equilibrium space can be non-linear. Due to the complexity of the data, there is a chance of error in the learned transformation. In order to remove the error in transformation, Error Removal using Correlation (ERC) technique is used. The description of the ERC technique is as follows:

Let the position of the test sample $\mathbf{t}^\chi$ in input space be $\mathbf{R}_{t, \chi}$ and after transformation, the position in equilibrium space $E$ be $\mathbf{R}_{t, E}$. In the first step, vector $\delta_i$ is added to $\mathbf{R}_{t, \chi}$ $\forall i$. In the next step, Spearman's rank correlation is computed between the transformed test point $\mathbf{R}_{t, E}$ in equilibrium space $E$ and the points obtained by adding vector $\delta_i$. The vector $\delta_i$ with maximum correlation value is used for final transformation of the test point $\mathbf{t}^\chi$.

\noindent \textbf{Stage 2: Coulomb Classification}\\
In order to classify a test sample, it is first transformed from input space to equilibrium space using the learned transformation from CNN as follows:
\begin{equation}
\mathbf{t}^E = \phi( \mathbf{t}^\chi,\mathbf{W},b)
\end{equation}
where, $\mathbf{t}^\chi$, $\mathbf{t}^E$ represents the test point in input and equilibrium space, respectively. During classification, a negative charge of unit magnitude $Q_t$ is assigned to the test sample $\mathbf{t}^E$. Each class in equilibrium space will attract the test sample towards itself. The test sample is assigned to the class that attracts the sample with maximum force. The value of constant $k$ in Coulomb's force equation is dependent on the medium. Similarly, in our case, the value of $k$ is dependent on the database and resolution which is obtained experimentally.

\begin{figure}[t]
\centering
\includegraphics[scale = 1.75]{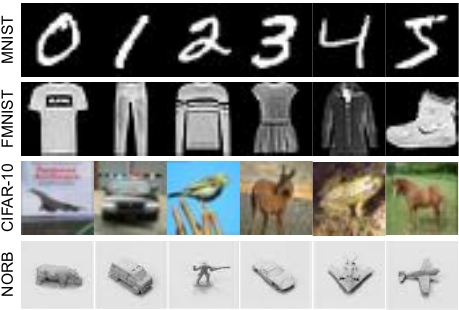}
\caption{Sample images of the MNIST, FMNIST, CIFAR-10, and NORB datasets.}
\label{fig:DB_Collage}
\end{figure}

\begin{figure*}[!h]
\centering
\includegraphics[scale = 1.8]{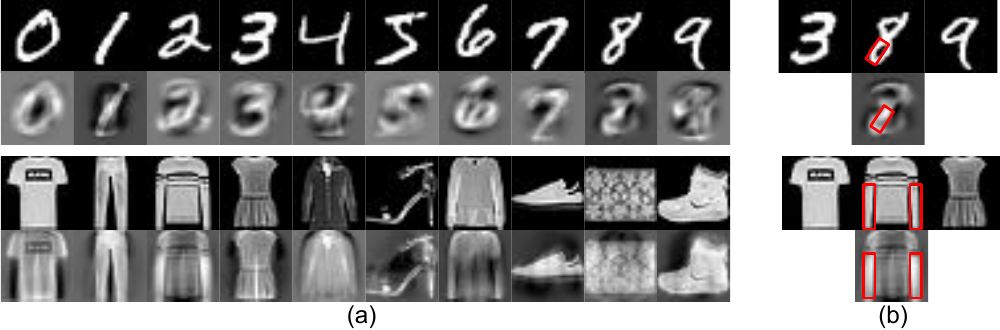}

\caption{Visualization of images in input and equilibrium space (a) samples from the MNIST and FMNIST  datasets. First row of each block correspond to the images in input space and the second row in the equilibrium space. (b) highlights the class distinguishing region in the images of the equilibrium space. For instance, the highlighted region of one class is not present in other classes.}
\label{fig:Equilibrium_visual}
\end{figure*}

\begin{table}[!b]
\centering
\caption{Value of $k$ used in the force equation for four datasets at three different resolutions.}
\label{k-value}
\renewcommand{\arraystretch}{1.2}
\begin{tabular}{|c|c|c|c|}
\hline
                       & \textbf{8x8} & \textbf{16x16} & \textbf{Original Resolution} \\ \hline
\textbf{CIFAR-10}      & 1            & 128            & 2048                         \\ \hline
\textbf{MNIST}         & 1            & 1              & 1                            \\ \hline
\textbf{FMNIST} & 1            & 1              & 1                            \\ \hline
\textbf{NORB}          & 1            & 1              & 512                          \\ \hline
\end{tabular}
\end{table}



\begin{table*}[h]
\centering
\caption{Computation time (in secs) taken by the proposed algorithm in computing and performing classification in equilibrium space.}
\label{ComputationTime}
\renewcommand{\arraystretch}{1.2}
\begin{tabular}{|l|c|c|c|c|c|c|}
\hline
\multirow{3}{*}{\textbf{Database}} & \multicolumn{3}{c|}{\textbf{Computation of equilibrium space}}                                          & \multicolumn{3}{c|}{\textbf{Classification}}                                                                                                      \\ \cline{2-7} 
                                   & \textbf{8x8} & \textbf{16x16} & \textbf{\begin{tabular}[c]{@{}c@{}}Original Resolution\end{tabular}} & \multicolumn{1}{c|}{\textbf{8x8}} & \multicolumn{1}{c|}{\textbf{16x16}} & \textbf{\begin{tabular}[c]{@{}c@{}}Original Resolution\end{tabular}} \\ \hline
\textbf{CIFAR-10}                  & 2.87         & 15.29          & 101.34                                                                  & 0.0015                            & 0.0021                              & 0.0047                                                                  \\ \hline
\textbf{MNIST}                     & 2.63         & 19.81          & 76.87                                                                   & 0.0016                            & 0.0018                              & 0.0061                                                                  \\ \hline
\textbf{FMNIST}             & 2.98         & 18.67          & 126.76                                                                  & 0.0018                            & 0.0019                              & 0.0059                                                                  \\ \hline
\textbf{NORB}                      &  3.12            &     19.24           &  367.23                                                                       &   0.0014                                & 0.0023                                    &  0.0127                                                                       \\ \hline
\end{tabular}
\end{table*}

\begin{figure*}[t]
\centering
\includegraphics[scale = 0.48]{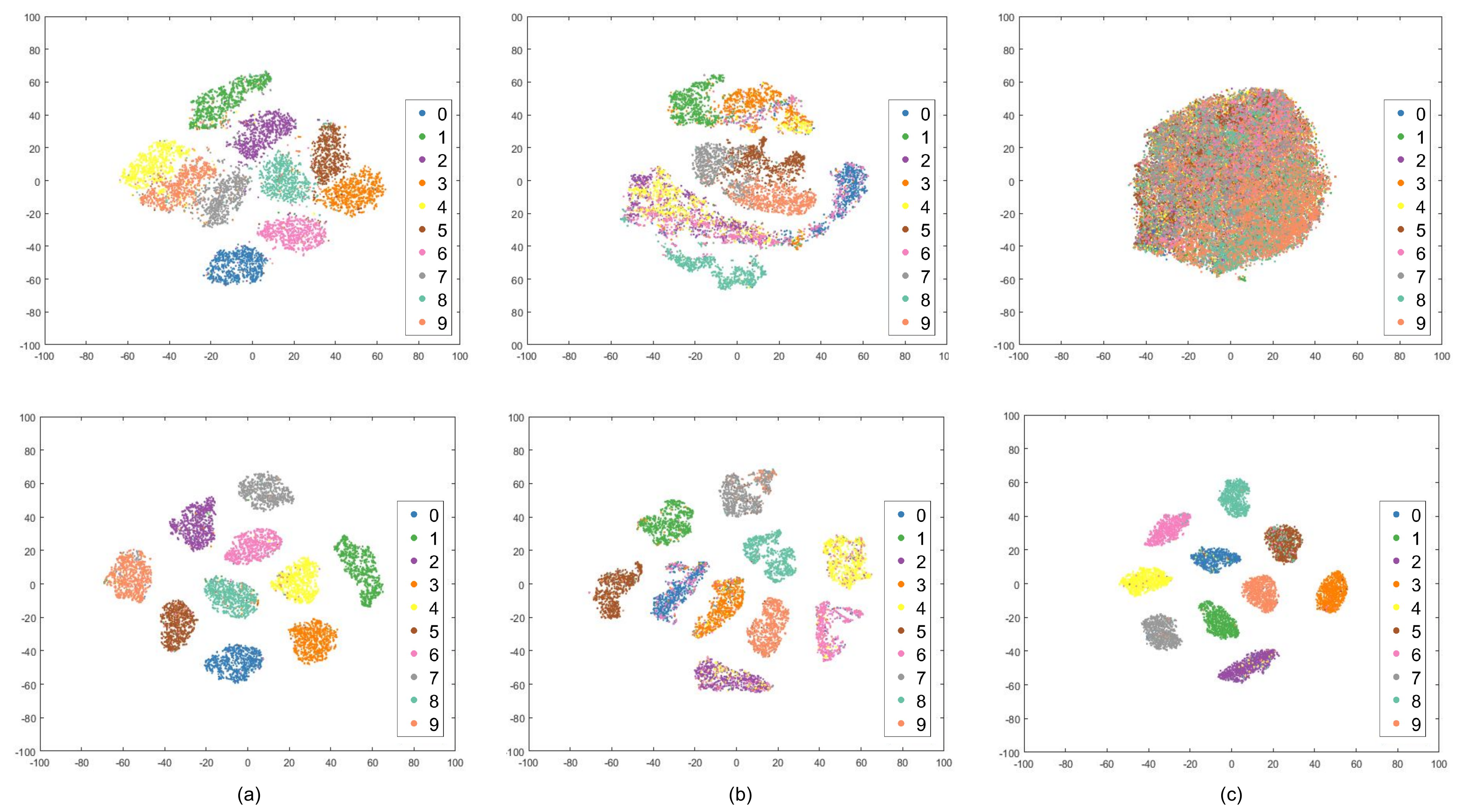}

\caption{t-SNE visualization of (a) MNIST, (b) FMNIST and (c) CIFAR-10 testing set at 8x8 resolution. The first row represents the visualization in input space and the second row in equilibrium space after classification.}
\label{fig:MNIST_FMNIST_CIFAR_TSNE}
\end{figure*}

\section{Experiments and Results}
In order to evaluate the performance of the proposed algorithm, experiments are performed on CIFAR-10 \cite{krizhevsky2009learning}, MNIST \cite{lecun1998mnist}, Fashion MNIST (FMNIST) \cite{xiao2017online}, and NORB \cite{lecun2004learning} datasets. Sample images of the datasets are shown in Fig. \ref{fig:DB_Collage}. Pre-defined protocols are used which ensure disjoint training and testing partitions. Following discuss the datasets detail.

\begin{itemize}
    \item \textbf{CIFAR-10} dataset contains 60,000, 32x32 color images of 10 different classes with 50,000 images in training set and 10,000 images in testing set.

    \item \textbf{MNIST} dataset consists of 60,000 training images and 10,000 testing images, of handwritten digits from 10 different classes (0-9). Each image is a grayscale image of 28x28 resolution. 
    
    \item \textbf{Fashion-MNIST} is a dataset of clothing and accessories with 60,000 images in training set and 10,000 images in testing set. Each image is a 28x28 grayscale image corresponding to one of the 10 class labels. 
    
    \item \textbf{NORB} dataset contains 96x96 grayscale images of toys in five different categories. The training set consists of 5 instances of each category while the testing set contains remaining 5 instances of each category. The dataset contains 24,300 images with each image being captured using two different cameras in training and testing sets. 
\end{itemize}

For each dataset, experiments are performed on 8x8, 16x16, and original resolution of images. In each experiment, equilibrium space is first computed followed by the classification in that space using Coulomb's law. Results and analysis of the experiments are discussed in sections \ref{Eq_Space_Compute} and \ref{Coulomb_Class}. Implementation details are given below.

\noindent \textbf{Implementation Details:} Experiments for computing equilibrium space are performed on MATLAB R2017a. The value of $k$ used in the force equation for each dataset and resolution is shown in Table \ref{k-value}. The deep learning model for learning the transformation between input space and equilibrium space comprises of an encoder-decoder architecture. The encoder architecture contains 3 blocks, each having (Conv$\rightarrow$ReLU$\rightarrow$Conv$\rightarrow$ReLU$\rightarrow$AvgPool) layers. Similarly, the decoder architecture has three blocks in the reverse order of encoder architecture. The model is trained for 30 epochs with a learning rate of 0.0001 for the first 20 epochs and reduced to 0.00001 for the next 10 epochs. A batch size of 64 is used with RMSprop optimizer. The code is implemented in tensorflow with 1080 Ti GPU.

\subsection{Computation of Equilibrium Space}
\label{Eq_Space_Compute}

Training set of each dataset is used for the computation of equilibrium space. Fig. \ref{fig:Intro_CIFAR_Image}(a) and \ref{fig:Intro_CIFAR_Image}(b) show the t-SNE visualization of CIFAR-10 training dataset in input and equilibrium space, respectively. Fig. \ref{fig:Intro_CIFAR_Image}(a) shows significant overlap between the classes in the input space. As mentioned in section \ref{Eq_Space}, Coulomb's force law is applied to the charges in input space so that the charges get settled in the equilibrium space by minimizing the overall force of the system. Data points of each class are then projected with respect to the position of each charge in the equilibrium space. The intra-class distance of each class in equilibrium space remains same as the input space. This results in the optimal separation of each class in the equilibrium space as illustrated in Fig. \ref{fig:Intro_CIFAR_Image}(b).

Fig. \ref{fig:Equilibrium_visual}(a) presents samples of MNIST and FMNIST dataset and the corresponding equilibrium representation. It is observed that the images in the equilibrium space contain regions of high intensity which correspond to the distinct features of the same class. For instance, in Fig. \ref{fig:Equilibrium_visual}(b), the lower left stroke of the number `8' is of high intensity, which is a distinguishing characteristic of the said class. Similarly, in the second example, the image corresponding to the full sleeves top contains high intensity regions at the sleeves. This demonstrates that the proposed algorithm is able to preserve discriminative features of a given class, which help in increasing the inter-class separability.


Table \ref{ComputationTime} shows that the proposed algorithm is computationally efficient in computing the equilibrium space. The computation time decreases as we decrease the resolution of the images.

\begin{table}[]
\centering
\caption{Classification accuracy (\%) of different algorithms on the CIFAR-10 dataset.}
\label{SOTA_Comp}
\renewcommand{\arraystretch}{1.2}
\begin{tabular}{|l|c|c|}
\hline
\textbf{Algorithms}            & \textbf{8x8}   & \textbf{16x16} \\ \hline
Partially Coupled Nets \cite{wang2016studying}        & 81.23          & -              \\ \hline
DenseNet \cite{huang2017densely}                      & 77.64          & -              \\ \hline
MobileNetV2 \cite{sandler2018mobilenetv2}                   & 77.72          & -              \\ \hline
EfficientNet \cite{tan2019efficientnet}                   & 73.88          & -              \\ \hline
RL-GAN \cite{xi2020see}                         & 88.11          & -              \\ \hline
DFD \cite{zhu2019low}                           & 81.26          & \textbf{90.41} \\ \hline
\textbf{Proposed (Before ERC)} & 85.70          & 86.71          \\ \hline
\textbf{Proposed (After ERC)}  & \textbf{96.26} & 87.08          \\ \hline
\end{tabular}
\end{table}


\subsection{Coulomb Classification}
\label{Coulomb_Class}

After computation of the equilibrium space, transformation between the input and equilibrium space is learned as discussed in section \ref{classification_eq}. Fig. \ref{fig:MNIST_FMNIST_CIFAR_TSNE} shows the t-SNE visualizations of the MNIST, FMNIST, and CIFAR-10 testing set at 8x8 resolution. It is observed that the classes are well separated in equilibrium space. The comparison of the proposed algorithm with existing methods on the CIFAR-10 dataset for low resolution images is shown in Table \ref{SOTA_Comp}. It is observed that the proposed algorithm achieves state-of-the-art performance at 8x8 resolution and comparable performance at 16x16 resolution. This shows the applicability of the proposed algorithm in real-world scenario for low resolution image classification.

Table \ref{Accuracy_Table} shows the classification performance of the proposed algorithm on 8x8, 16x16 and original resolution for all the datasets. It is observed that the ERC technique enhances the performance in most of the cases. For instance, the accuracy on the FMNIST dataset at 8x8 resolution after ERC is 87.30\% which is approximately 10\% better than before applying ERC. It is interesting to observe that, in some cases, the performance at 8x8 resolution is better than 16x16 and original resolution. For instance, the classification accuracy of the CIFAR-10 dataset at 8x8, 16x16, and original resolution is 96.26\%, 87.08, and 91.06, respectively. It is our assertion that, at low resolution, the deep model is able to learn the transformation better than at higher resolution.

\begin{table}[t]
\centering
\caption{Classification accuracy (\%) of the proposed algorithm on four different datasets at three different resolutions.}
\label{Accuracy_Table}
\renewcommand{\arraystretch}{1.2}
\begin{tabular}{|l|c|c|c|c|c|c|}
\hline
\multirow{3}{*}{\textbf{Database}} & \multicolumn{2}{c|}{\textbf{8x8}}                                          & \multicolumn{2}{c|}{\textbf{16x16}}    &   \multicolumn{2}{c|}{\textbf{Original Resolution}}                                                                                                    \\ \cline{2-7} 
                                   & \textbf{\begin{tabular}[c]{@{}c@{}}Before  \\ ERC\end{tabular}} & \textbf{\begin{tabular}[c]{@{}c@{}}After  \\ ERC\end{tabular}} & \textbf{\begin{tabular}[c]{@{}c@{}}Before  \\ ERC\end{tabular}} & \textbf{\begin{tabular}[c]{@{}c@{}}After  \\ ERC\end{tabular}}& \textbf{\begin{tabular}[c]{@{}c@{}}Before  \\ ERC\end{tabular}}& \textbf{\begin{tabular}[c]{@{}c@{}}After \\ ERC\end{tabular}} \\ \hline
CIFAR-10                           & 85.70                                                          & 96.26                                                         & 86.71     & 87.08                                                                                                          & 90.13                                                          & 91.06                                                                                                       \\ \hline
MNIST                              & 96.29                                                          & 96.21                                                         & 98.44                                                          & 98.37                                                    & 99.23                                                          & 99.23                                              \\ \hline
FMNIST                      & 77.25                                                          & 87.30                                                         & 84.39                                                          & 84.79                                                        & 91.33                                                          & 91.44                                                                       \\ \hline
NORB                               & 82.30                                                          & 84.30                                                         & 83.18                                                          & 84.45                                                 & 93.41                                                          &   93.27                                                       \\ \hline
\end{tabular}
\end{table}





The computation time to transform data points from input space to equilibrium space is dependent on the model used for learning the transformation. The time taken by the proposed algorithm for classification in equilibrium space is shown in Table \ref{ComputationTime}. It is observed that the time taken for classification is significantly low.

\section{Conclusion}
In machine learning, the aim of the majority of projection algorithms and loss functions is to increase the inter-class distance. However, increasing the inter-class distance can affect the intra-class distance. Maintaining a balance between the two is a challenging task. This paper proposes a new algorithm to compute the equilibrium space where the inter-class separation among the classes is optimal without affecting the intra-class distance of the data distribution. The algorithm further learns the transformation from input space to equilibrium space to perform classification in equilibrium space. Also, an Error Removal using Correlation (ERC) technique is used to enhance the performance. Experiments are performed on four different datasets. For each dataset, experiments are performed on three different resolutions of images. It is found that the proposed algorithm achieves state-of-the-art result on the CIFAR-10 dataset at 8x8 resolution. In the future, we plan to extend the application of the proposed algorithm in cross-domain matching scenarios such as matching low resolution faces with high resolution images \cite{bhattTIP} or sketch to photo matching \cite{nagpal2021discriminative}.    

\section*{Acknowledgements}
Puspita Majumdar is partly supported by DST INSPIRE Ph.D. Fellowship. Mayank Vatsa is partially supported through Swarnajayanti Fellowship by the Government of India.

\bibliographystyle{abbrv}
\bibliography{IEEEexample}

\end{document}